\documentclass[letterpaper, 10 pt, conference,dvipsnames]{ieeeconf}  

\IEEEoverridecommandlockouts                              
\usepackage[utf8]{inputenc}
\usepackage[T1]{fontenc}
 \usepackage[export]{adjustbox} 
 \usepackage{diagbox}
\usepackage{colortbl}

\usepackage{color}
\usepackage{xcolor}
\usepackage{commath}
\usepackage{tabularx}
\usepackage{diagbox} 
\usepackage{cancel}
\usepackage{bm}

\usepackage{graphicx} 
\usepackage{amsmath} 
\usepackage{amssymb}  
\usepackage{cancel}
\usepackage{verbatim}
\usepackage{algorithm}
\usepackage{algpseudocode}
\usepackage{mathtools}
\usepackage{float}
\usepackage{tikz}
\usepackage{comment}
\usepackage{subcaption}
\usepackage{hyperref}
\usepackage{siunitx}
\usepackage{soul}

\usetikzlibrary{automata,positioning}
\tikzstyle{node}=[align=center]

\let\labelindent\relax
\usepackage[inline]{enumitem}

\usepackage{comment}
\usepackage{xcolor}
\usepackage{xspace}
\usepackage[normalem]{ulem}

\definecolor{OliveGreen}{RGB}{0,200,25}
\newcommand{\red}[1]{\textcolor{red}{#1}}

\newcommand{\darkgreen}[1]{\textcolor{OliveGreen}{#1}}
\newcommand{\blue}[1]{\textcolor{blue}{#1}}
\newcommand{\orange}[1]{\textcolor{orange}{#1}}

\newcommand{\ie}{i.\,e.,\xspace}
\newcommand{\eg}{e.\,g.,\xspace}

\newcommand{\replaced}[2]{\red{\ifmmode\text{\sout{\ensuremath{#1}}}\else\sout{#1}\fi}\;\darkgreen{#2}}
\newcommand{\removed}[1]{\red{\ifmmode\text{\sout{\ensuremath{#1}}}\else\sout{#1}\fi}}

\newcommand{\remark}[1]{\blue{--- #1 ---}}
\newcommand{\todo}[1]{\{\orange{---TODO--- #1}\}}
\newcommand{\toremove}[1]{#1}

\newif\iffinal

\iffinal
	
	\renewcommand{\replaced}[2]{#2}
	\renewcommand{\removed}[1]{}
	\renewcommand{\remark}[1]{}
	\renewcommand{\todo}[1]{}
	\renewcommand{\toremove}[1]{}
\fi

\newcommand{\removedfootnote}[1]{\footnote{\removed{#1}}}
\newcommand{\removedsubsection}[1]{\subsection{\texorpdfstring{\removed{#1}}{#1}}}

\iffinal
	
	\renewcommand{\removedfootnote}[1]{}
	\renewcommand{\removedsubsection}[1]{}
	
	\renewcommand{\removedsubsection}[1]{}

\fi

\title{\LARGE \bf

Uncertainty-aware Risk Assessment of Robotic Systems via \\ Importance Sampling
}

\author{Woo-Jeong Baek, Tom P. Huck, Joschka Haas, Jonas Lewandrowski, Tamim Asfour, and Torsten Kröger 
\thanks{The authors are with the Institute for Anthropomatics and Robotics (IAR), Karlsruhe Institute of Technology (KIT)
 {\tt\small \{baek, tom.huck, asfour, torsten\}@kit.edu}.\newline
This work was funded by the German Federal Ministry for Economic Affairs and Climate Action under the project ``SDM4FZI'' (\url{www.sdm4fzi.de})}
}

\makeatletter
\def\namedlabel#1#2{\begingroup
    #2%
    \def\@currentlabel{#2}%
    \phantomsection\label{#1}\endgroup
}
\makeatother
\begin{document}

\maketitle
\thispagestyle{empty}
\pagestyle{empty}

\begin{abstract}


In this paper, we introduce a probabilistic approach to risk assessment of robot systems by focusing on the impact of uncertainties. 
While various approaches to identifying systematic hazards (\eg bugs, design flaws, etc.) can be found in current literature, little attention has been devoted to evaluating risks in robot systems in a probabilistic manner. 
Existing methods rely on discrete notions for dangerous events and assume that the consequences of these can be described by simple logical operations. 
In this work, we consider measurement uncertainties as one main contributor to the evolvement of risks.
Specifically, we study the impact of temporal and spatial uncertainties on the occurrence probability of dangerous failures, thereby deriving an approach for an uncertainty-aware risk assessment. 
Secondly, we introduce a method to improve the statistical significance of our results: 
While the rare occurrence of hazardous events makes it challenging to draw conclusions with reliable accuracy, we show that importance sampling -- a technique that successively generates samples in regions with sparse probability densities -- allows for overcoming this issue. 
We demonstrate the validity of our novel uncertainty-aware risk assessment method in three simulation scenarios from the domain of human-robot collaboration. 
Finally, we show how the results can be used to evaluate arbitrary safety limits of robot systems. 
\end{abstract}

\section{Introduction} \label{introduction}
Robots which are deployed in safety-critical applications such as physical human-robot collaboration need to be analyzed thoroughly to determine if they pose any unacceptable risk to human safety. In such analyses, there are two distinct classes of risks to consider. Risks can result from systematic errors. This includes, for instance, software bugs, inappropriate choice of safety measures, or other design flaws. However, risks can also result from probabilistic effects, such as random hardware failures or measurement uncertainties. While there are several approaches to identifying systematic errors \cite{Leveson2016,Clarke2018,Huck2020}, there is still comparatively little work on how to consider probabilistic effects and uncertainties in the risk assessment. Traditionally, safety engineers rely on methods such as reliability block diagrams (RBD) or fault tree analysis (FTA) for probabilistic analyses \cite{Rausand2021}. However, as will be discussed below, these methods rely on restrictive assumptions which limit their applicability in a robotics context. In this paper, we argue for stronger use of probabilistic representations in the risk assessment of robotic systems. In particular, we study how measurement uncertainties contribute to dangerous situations. 
Here, measurement uncertainties reflect possible deviations of a parameter due to the limited accuracy in the measurement process. 
Such inaccuracies are caused by technical limitations of system components or environmental disturbances. 
In fact, the existence of measurement uncertainties motivates the probabilistic representations of parameters. 
This means that parameters are not assigned to scalar values, but probability density functions instead. 
In this contribution, we develop an uncertainty-aware risk assessment method by accounting for temporal and spatial measurement uncertainties. 
To do so, we model three human-robot collaboration (HRC) scenarios in the simulation environment \emph{CoppeliaSim} and explore how temporal and spatial uncertainties accumulate over the robot system and contribute to the risk evolvement. In the next step, we extend our investigation of risk sources by employing importance sampling. 
Briefly, this statistical method allows to derive conclusive inferences regarding the origins of dangerous events despite their rare occurrence probability. 
By presenting our novel method on assessing risks of robot systems in a probabilistic manner, we describe how our approach can contribute to evaluating arbitrary safety limits.

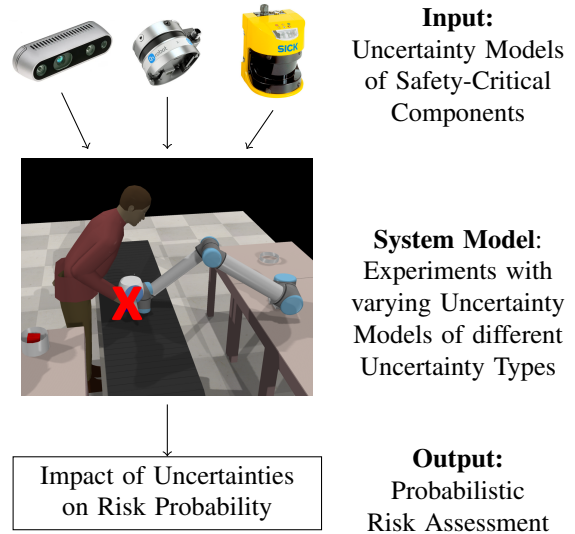
\begin{figure}[t!]
	\centering
		\begin{tikzpicture}

           %
        \node[text width=4em, align=center, node distance=0em] (LS_SICK) {\includegraphics[trim=0 0 0 0, clip, width=3em]{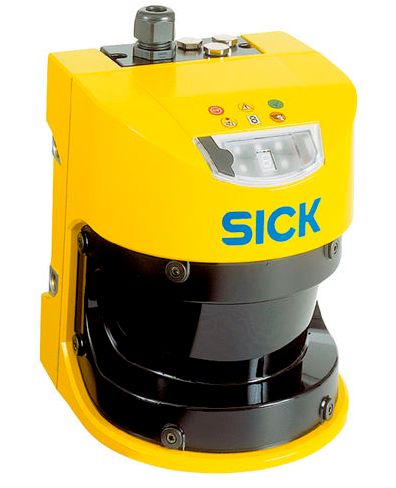}};
        
        \node[text width=4em, align=center, left of=LS_SICK, node distance=2em, xshift = -2em] (FTS) {\includegraphics[trim=0 0 0 0, clip, width=3em]{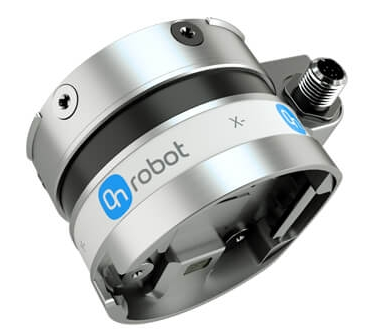} };

            \node[text width=4em, align=center, left of=FTS, node distance=2em, xshift = -2em] (intelRealSense) {\includegraphics[trim=0 0 0 0, clip, width=4em]{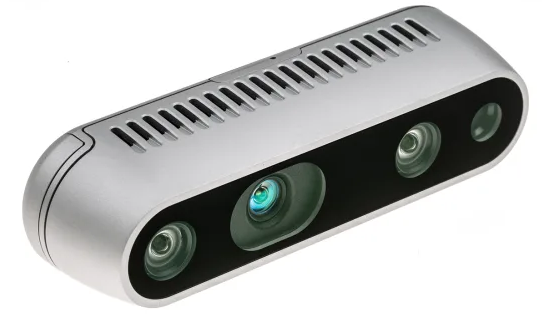} };

            \node[text width=12em, align=center, below of=FTS, node distance=8.5em] (sim_coll) {\includegraphics[trim=0 0 0 0, clip, width=11em]{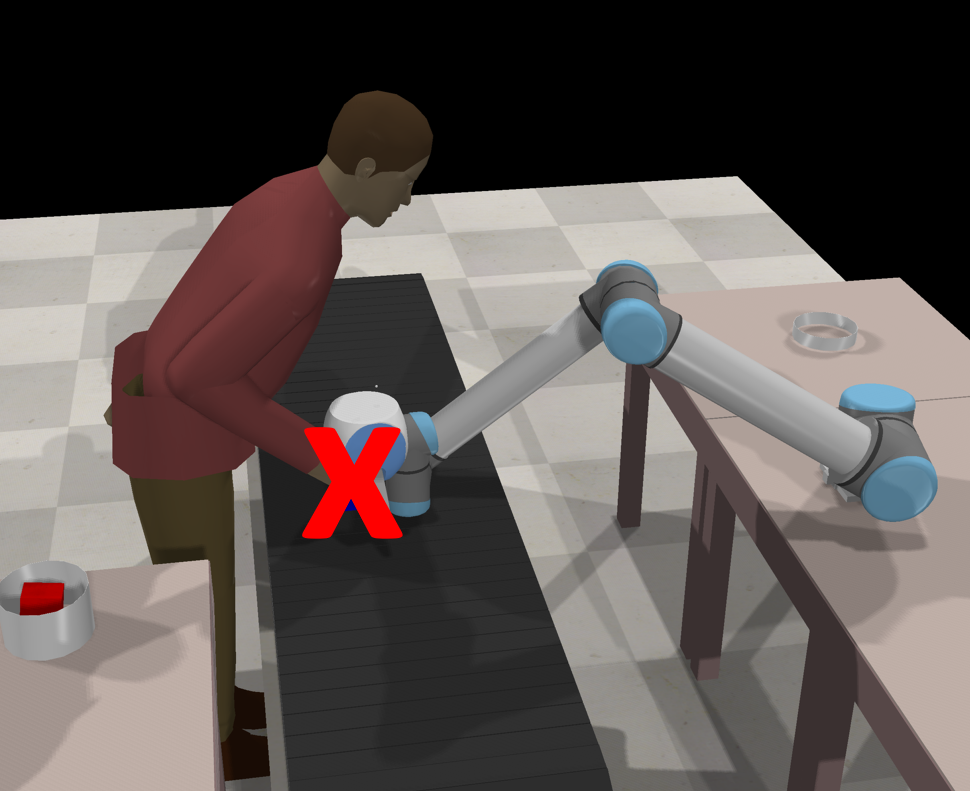} };
        
            \node[draw, text width=11em, align=center, below of=sim_coll, node distance=8.2em] (Risk) {Impact of Uncertainties on Risk Probability};







            \node[text width=9em, align=center, yshift= -0.2cm, right of=LS_SICK, node distance=7em] (text1) {\textbf{Input:}\\ Uncertainty Models of Safety-Critical Components };
            \node[text width=9em, align=center, below of=text1, node distance=9em] (text2) {\textbf{System Model}:\\ Experiments with varying Uncertainty Models of different Uncertainty Types};
            \node[text width=9em, align=center, yshift = 0.7cm, below of=text2, node distance=9em] (text3) {\textbf{Output:}\\Probabilistic Risk Assessment};
   
             \path (sim_coll.north)+(-3:0) coordinate (LS_goal);
			 
            \draw[->] (LS_SICK.south) -> ([xshift=3em] sim_coll.north);
           \draw[->] (intelRealSense.south) -> ([xshift=-3em] sim_coll.north);
           \draw[->] (FTS.south) -> node [text width=2cm,midway,right=0.3em,yshift=0.7cm ] { } (sim_coll.north);
            \draw[->] (sim_coll.south) -> node [text width=2cm,midway,right=0.3em,yshift=0.7cm ] { } (Risk.north);
         

		\end{tikzpicture}
	\caption{We develop an uncertainty-aware risk assessment method for robotic systems.} 
	\label{fig:Fig1}
\end{figure}



\section{Related Work} \label{sota}
\subsection{Probabilistic Risk Assessment Methods}
Fault Tree Analysis (FTA) and Reliability Block Diagrams (RBD) are graph-based probabilistic risk assessment methods \cite{Hasan2015,Ruijters2015}. FTA is based on a tree graph where the root is an undesired event (\eg an accident), and nodes/leafs are events contributing to it (\eg component failures). Nodes are connected by logical gates (\textit{AND, OR}). Probabilities are assigned to basic events and propagated to the top via these gates. RBDs are block diagrams where each block represents a critical system function. Series-connected blocks depend on each other, while parallel blocks are redundant. The analysis is performed similarly to FTA with boolean logic. FTA and RBD are dual representations and can be converted into each other.
FTA and RBD are limited in several ways:  They only consider events with binary outcomes and fixed probabilities. Events must be statistically independent. Temporal aspects like the order in which events occur are not considered \cite{Rausand2021}.
Dynamic fault trees (DFT) \cite{NASA2002} and Bayesian Networks (BN) \cite{Weber2012} relax some of these limitations. DFT are extensions of FTA for dynamic (i.e., time-dependant) failure scenarios \cite{NASA2002}. Yet, DFTs only handle certain classes of dynamic behavior and are still relatively inflexible.
Bayesian networks (BN) also rely on graph-based representations, yet their structure is more flexible and not bound to a tree structure. States are not restricted to binary values, and independence assumptions are relaxed \cite{Sucar2015}. 
Markov chains describe systems in terms of probabilistic state transitions and are thus well-suited for modeling dynamic effects in probabilistic risk analyses \cite{Rausand2021}. Markov models can be evaluated for instance by probabilistic model checking \cite{Zhao2019} or numerically \cite{Faghih2014}. 
The Monte Carlo method (MC) is also commonly used for risk analyses \cite{stroeve2009,Arunraj2013}. Here, one samples the input space of a system model, computes a model output for each sample, and observes the resulting output distributions \cite{Rausand2021}. Since the model is treated as a black box, MC is suitable for analytically intractable methods. For instance, MC is used to evaluate complex DFTs \cite{Rao2009,Gascard2018,Kubo2022}. However, the accuracy of MC is limited when the events of particular interest are rare in the input distributions. In such cases, a large number of samples is required to accurately assess the effects of such rare events. This may be infeasible for evaluating computationally expensive system models. Rare-event simulations such as importance sampling address this limitation \cite{Rubino2009}. Although some works propose rare-event techniques for probabilistic risk assessment \cite{Hu2022,Arief2021,OKelly2018,Puch2018}, the field is still sparsely researched, especially in a robotics context.


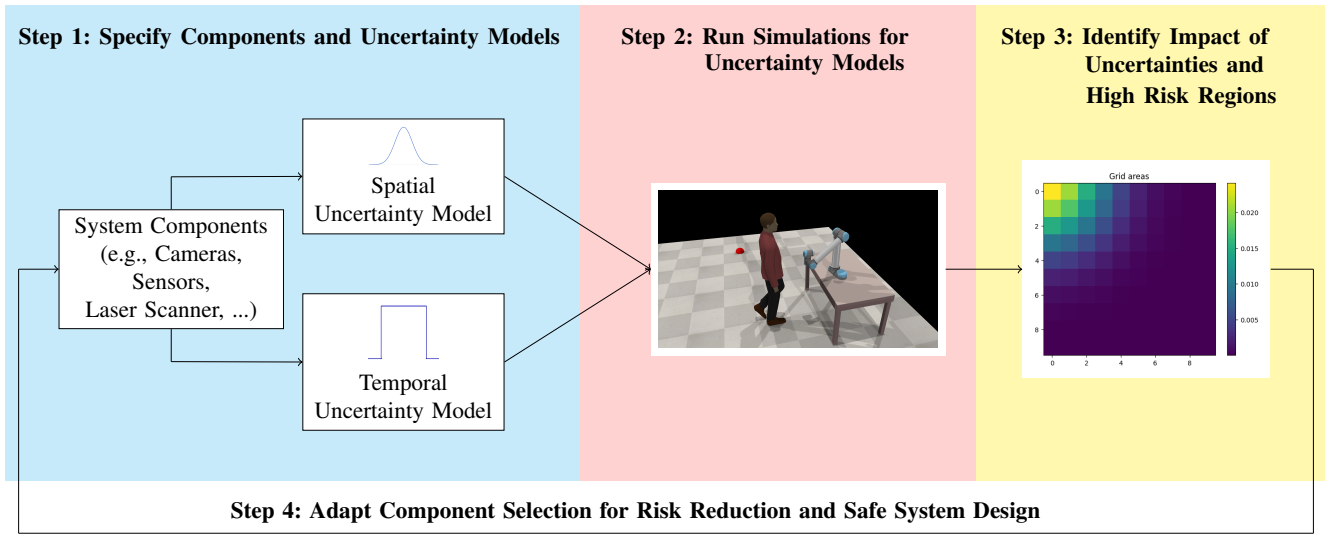
\begin{figure*}[t!]
	\centering
    \resizebox{1\textwidth}{!}{
		\begin{tikzpicture}
        \fill[cyan!20] (-12,-3.2) rectangle (-3.3,4); \node at (-7.7,3.5) {\textbf{Step 1: Specify Components and Uncertainty Models}};
        
        \fill[pink!70] (-3.3,-3.2) rectangle (2.7,4);  \node at (-0.5,3.5) {\textbf{Step 2: Run Simulations for}};
                                                    \node at (0.1,3.1) {\textbf{Uncertainty Models}};
        
        \fill[yellow!40] (2.7,-3.2) rectangle (8,4); \node at (5.1,3.5) {\textbf{Step 3: Identify Impact of }};
                                                    \node at (5.7, 3.1) {\textbf{Uncertainties and}};
                                                     \node at (5.8, 2.6) {\textbf{High Risk Regions }};

        \node[fill=white, text width=12em, align=center, node distance=0em] (simulation) {\includegraphics[trim=0 0 0 0, clip, width=12em]{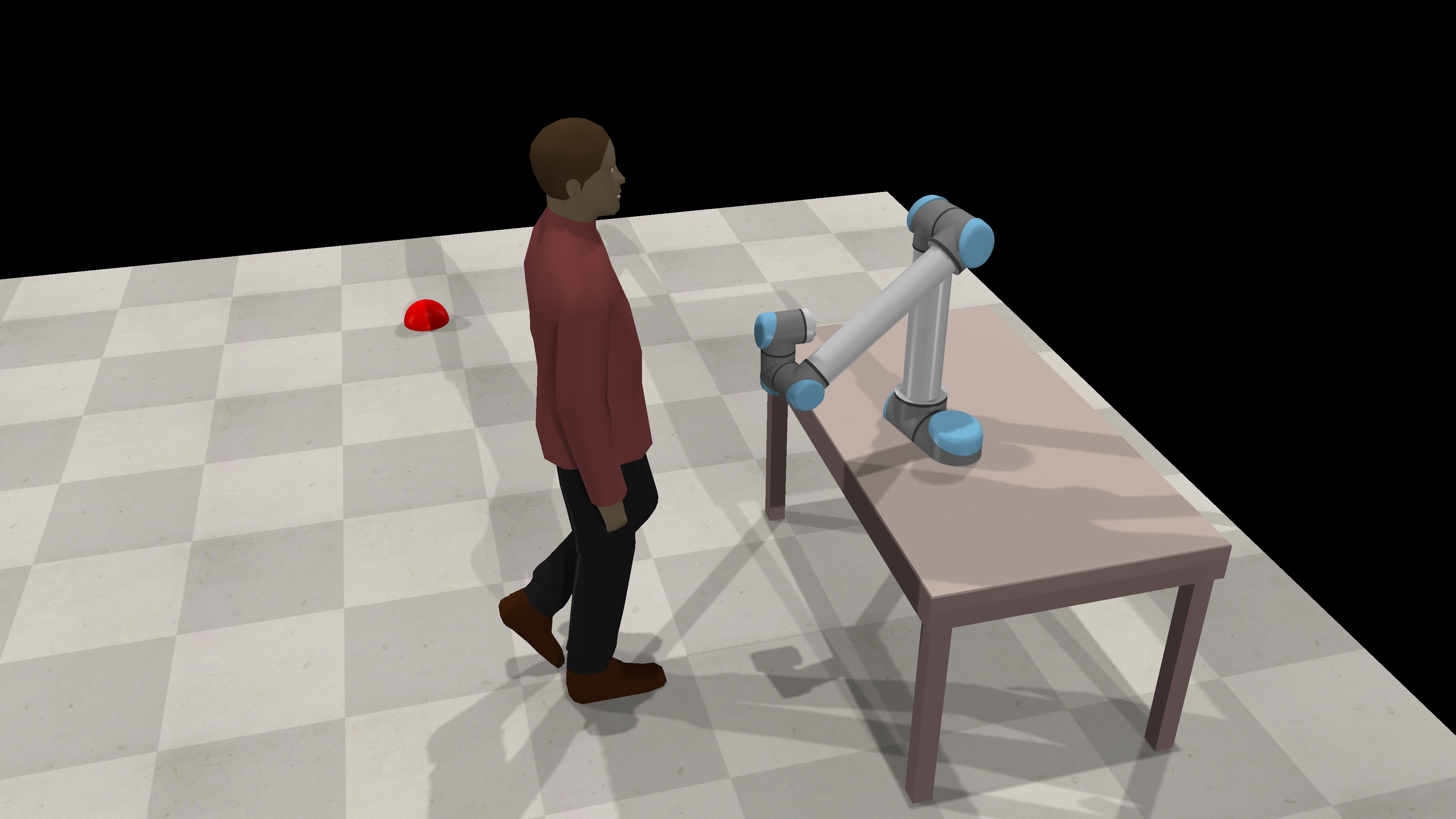}};

            \node[draw, fill=white, text width=8em, align=center, left of=simulation, node distance=17em, yshift= -4em] (uncertainty1) {\includegraphics[trim=0 0 0 0, clip, width=3em]{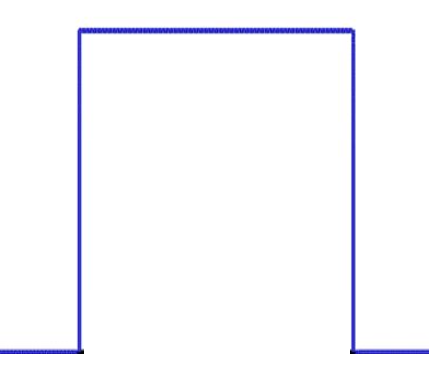} \\ Temporal \\ Uncertainty Model};

            \node[draw, fill=white, text width=8em, align=center, above of=uncertainty1, node distance=8em] (uncertainty2) {\includegraphics[trim=0 0 0 0, clip, width=3em]{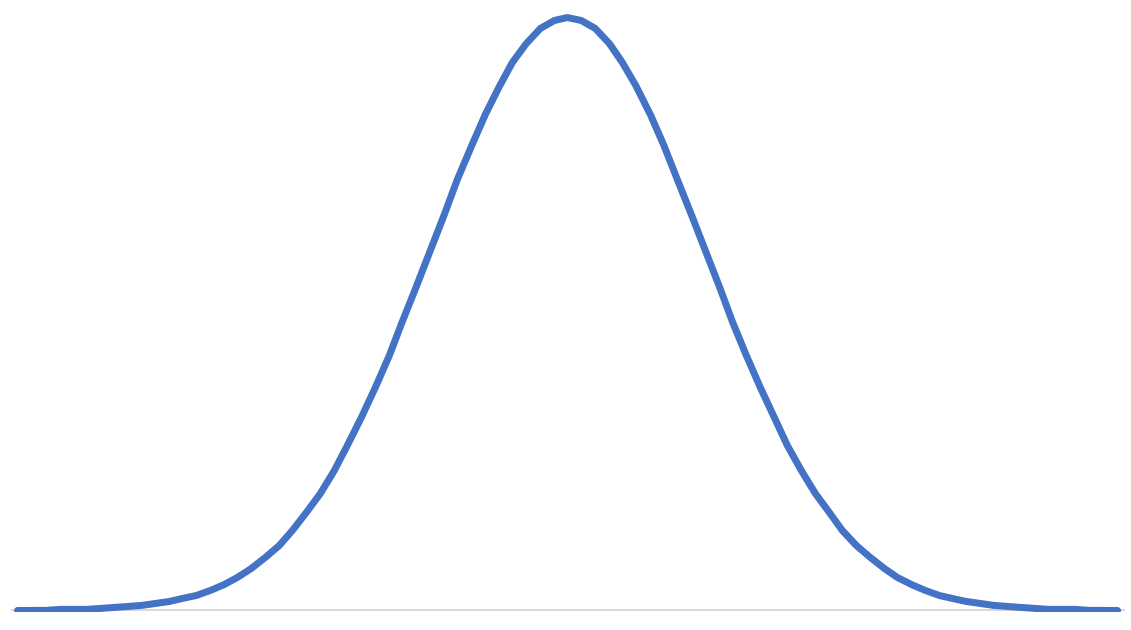} \\ Spatial \\ Uncertainty Model};



            \node[draw, fill=white, text width=9em, align=center, left of=uncertainty1, node distance=10em, yshift = 4em] (Components) { System Components \\(e.g., Cameras, Sensors, \\ Laser Scanner, ...)};

            \node[ fill=white, text width=10em, align=center, right of=simulation, node distance=15em] (risk) {\includegraphics[trim=0 0 0 0, clip, width=10em]{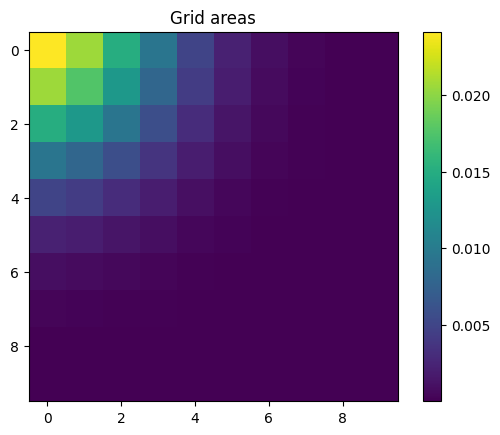} };




            \node[text width=40em, align=center, below of=simulation, node distance=10.5em, xshift = -7em] (text4) {\textbf{Step 4: Adapt Component Selection for Risk Reduction and Safe System Design}};

            \draw[->] (uncertainty1.east) -> node [text width=2cm,midway,right=0.3em,yshift=0.7cm ] { } (simulation.west);
            \draw[->] (uncertainty2.east) -> node [text width=2cm,midway,right=0.3em,yshift=0.7cm ] { } (simulation.west);
            \draw[->] (simulation.east) -> node [text width=2cm,midway,right=0.3em,yshift=0.7cm ] { } (risk.west);

            \draw[-] (risk.east) -- (7.8,0);
            \draw[-] (7.8,0) -- (7.8,-4);
            \draw[-] (7.8,-4) -- (-11.8,-4);
            \draw[-] (-11.8,-4) -- (-11.8,0);
            \draw[->] (-11.8,0) -- (Components.west);
        
            \draw[-] (Components.north) -- (-9.49, 1.4);
            \draw[-] (Components.south) -- (-9.49, -1.4);
            \draw[->] (-9.49,1.4) -- (uncertainty2.west);
            \draw[->] (-9.49,-1.4) -- (uncertainty1.west);

		\end{tikzpicture}
  }
	\caption{We present an uncertainty-aware risk assessment method by feeding samples from different uncertainty models into the simulator. The obtained results are analyzed with respect to the probability of the occurrence of dangerous events. In addition, we introduce grid-based importance sampling to draw conclusions with sufficient statistical significance.}
	\label{fig:overview_risk_guided_search}
\end{figure*}

\subsection{Uncertainty Propagation in Robotics}
In this paper, we rely on the uncertainty notions according to the \emph{Guide to the Evaluation of Uncertainty in Measurement (GUM)} in \cite{GUM2009}. 
While incorporating measurement uncertainties in the risk assessment of robot systems has not been thoroughly studied to date, works from the field of mobile robot navigation show methodological similarities. 
For example, the contributions in \cite{Luo2020} and \cite{Arvanitakis2017} implement probabilistic approaches in the control algorithms. 
Specifically, the authors in \cite{Luo2020} derive probabilistic safety barrier certificates based on Gaussian distributions. 
These certificates define the robot control space and guarantee that safety conditions are held. 
Contrarily, the authors in \cite{Arvanitakis2017} alleviate inaccurate navigation behavior by means of an uncertainty space.
Although a control law with promising performance is developed, this work suffers from the assumption that the uncertainty is known beforehand. 
Furthermore, the authors in \cite{Roy1999} and \cite{Giancola2018} show how sensor uncertainties can be taken into account. 
In \cite{Roy1999}, the authors propose a method based on the Bayes formulation to infer sensor data with the robot position. 
In particular, uncertainties of sensors are used to perform a Markov localization. 
In contrast, the authors in \cite{Giancola2018}  study the uncertainties of three camera types (Time of Flight, Structured Light and Active Stereo) in the face of environmental disturbances, thereby discussing the influence of various parameters. 
Apart from these works, contributions as \cite{Riaz2020}, \cite{Loquercio2020} and \cite{Zhang2023} use learning methods to account for uncertainties and their propagation behavior. 
Particularly, the work in \cite{Zhang2023} introduces a safety-critical control framework by modeling uncertainties with a radial basis function neural network (RBFNN). 
Here, the input data consists of the robot control force, the angular position and velocity of the robot.  
By modeling the weights with radial basis functions, negative effects due to uncertainties and their propagation are captured. 
While this may help to circumvent negative consequences due to uncertainties, their impact is not explored in the context of safety compliance in HRC.

\subsection{Importance Sampling in Robot Systems}
Several works have shown that importance sampling contributes to the analysis of interesting regions in robot systems. 
For example, the authors in \cite{OKelly2018} present a work on autonomous vehicle testing with the goal of developing safe systems for real-world environments. 
In fact, an algorithm for adaptive importance sampling is considered to overcome the issue of the rare occurrence of accidents. 
The authors characterize the search space by a safety parameter and use a cross-entropy method to approximate the optimal importance distribution. 
After identifying suitable parameters with a physics-based simulator, autonomous vehicle testing is performed on real-world data. The findings demonstrate significant speedups over existing real-world testing methods. 
On the other hand, the work in \cite{Kurniawati2004} addresses a roadmap planning problem of robots.  
To develop a method that performs reliably in narrow passages, importance sampling is employed to overcome the problem of limited samples in interesting areas. 
The underlying idea of the work is that narrow passages in the configuration space can be identified by narrow passages in the workspace. Thus, the authors in \cite{Kurniawati2004} model the workspace as a polyhedron, and conduct the sampling process for a fixed resolution to assign importance values to the sampled points. 
The results show that free configurations can be obtained with enhanced performance by employing importance sampling. 
Similarly, the authors in \cite{Luo2019} introduce a general framework IS-DESPOT for robot planning by considering uncertainties. 
First, offline simulations are conducted. 
After searching for an optimal plan in a set of sampled scenarios to learn the important distributions, these are used for online planning purposes. 

\section{Preliminaries}
In the following, we introduce the terminology and tools we use in this paper. 
\subsection{Measurement Uncertainties}\label{subsec:MeasurementUncertainties}
As mentioned above, we rely on the metrological viewpoint of uncertainties in \emph{GUM} in \cite{GUM2009}. 
Accordingly, measurement uncertainties reflect the missing knowledge on parameters caused by inaccuracies in the measurement process. 
Possible sources are given by technical limitations of tools, unexpected environmental disturbances and the lack of data. 
In particular, measurement uncertainties are distinguished in statistical uncertainties $u_{stat}$ and systematic uncertainties $u_{sys}$. 
While the latter ones denote tool-specific uncertainties of system components that are usually specified in the data sheets, $u_{stat}$ draws from the limited availability of data. 
In addition, measurement uncertainties are characterized with respect to an \emph{attribute a}. 
Practically, $a$ specifies for which parameter the measurement uncertainty is calculated. 
In this work, we focus on the measurement uncertainties of safety-critical parameters in HRC. 
By doing so, spatial and temporal uncertainties, denoted by the $u_s$ and $u_t$, respectively, are studied. 
We refer to the measurement uncertainty by the term \emph{uncertainty} for the remainder of this work. 
Furthermore, the uncertainty of an entire system, that results from the propagation of uncertainties of single components, is denoted with $u_{prop}$. 
To obtain $u_{prop}$, we make use of the Monte-Carlo (MC) sampling method presented in our previous contribution \cite{Baek2023_ICRA}. 

\subsection{Importance Sampling} \label{subsec:IS}
The general purpose of statistical sampling lies in drawing individual data points from a population that is represented by a probability density function (PDF), denoted with $p(x)$ for a random variable $x$.
Logically, a high number of samples is generated in areas with high probabilities, while only few samples are drawn from regions with sparse probabilities. 
However, this poses a challenge for deriving conclusive results for these regions due to the limited number of samples, and thus a high statistical uncertainty $u_{stat}$. 
Particularly, calculating the expectation value $\mathbb{E}(x)$ via MC sampling is bound to a high variance in the approximation error (VAE). 
To overcome this, importance sampling suggests shifting the PDF into the region with a low probability density. 
In fact, a separate, latent PDF $q(x)$ is generated and taken into account for estimating the expectation value, that is: 
\begin{equation}
\begin{split}
\mathbb{E}(x) &  \coloneqq \int f(x) p(x) dx = \int f(x) \frac{p(x)}{q(x)} q(x) dx\\
      & \coloneqq \int f(x) \omega(x) q(x) dx,
\end{split}
\label{eq:GBIS}
\end{equation}
where $q(x) \neq 0$ and $\omega(x)$ is the sampling weight. 
By selecting the latent distribution $q(x)$ in regions where $|f(x)|\cdot p(x)$ is large, the VAE for calculating $\mathbb{E}(x)$ can be significantly reduced. We refer the reader to \cite{Biondini2015} for details.

\section{Problem Statement and Motivation}
Generally, our goal is to probabilistically evaluate the risk that is defined in standard ISO 12100 by
\begin{equation}
    risk(i) = r(severity (i), Pr(i)). 
    \label{eq:risk}
\end{equation}
Here, $Pr(i)$ denotes the probability for the occurrence of incident $i$. 
We view measurement uncertainties as significant contributors to $Pr(i)$, thereby motivating dedicated analyses regarding the impact of $u_t(a_c)$ and $u_s(a_c)$ on the risk probability. 
To be specific, we seek to identify a mapping
\begin{equation}
    u_{prop}(a_c) = m(u_t(a_c), u_s(a_c)) \mapsto Pr(i),
    \label{eq:mapping}
\end{equation}
where $u_{prop}(a_c)$ reflects the propagated uncertainty $u_{prop}$ arising from the entire application regarding the critical attribute $a_c$. 
The functional relationship $m(\cdot)$ describes the relationship between $u_t$ and $u_s$ that must be identified individually for each system. \\
Secondly, we aim at analyzing dangerous situations in more detail. 
Apart from the influence of uncertainties, further system characteristics may contribute to the evolvement of risks. 
However, the rare occurrence of dangerous situations makes it difficult to properly investigate their origins. 
As explained in \autoref{subsec:IS}, importance sampling (IS) provides the possibility of enhancing the probability density in specific regions. 
To fully leverage the advantages of IS, we need to identify an appropriate importance distribution $q(x)$ that effectively reduces the VAE. 
In this contribution, we focus on grid-based IS: Here, the workspace is partitioned into discrete grids within the sample space $\mathcal{D}$. To obtain an importance distribution that implements an occupancy grid, we draw statistically independent samples $x_1,...x_n$ with $n \in \mathbb{N}$ generated from the sample space $X$ under $p(x)$. In addition, we define a grid partition $G$ within $\mathcal{D}$. 
With $g \in G$ and $g_x = \{x_i | x_i \in g\}$ where the number of grids is given by $G_x$. 
Consequently, each grid is assigned to a probability.
Hence, our objective is to derive the ideal number for $G_x$ such that the importance grid function 
\begin{equation}
    q_G(x)=\frac{\sum_{x \in G_x}A_g}{\sum_{i=1}^n x_i},
\end{equation}
where $A_g$ denotes the area of cell $g$. 
Thus, the density is obtained via calculating the fraction between the cell of one area and the area of the entire sampling space. 
falls in the region, where $|f(x)|\cdot p(x)$ is large.

\section{Approach} 
In the following, we introduce our approach by referring to \autoref{fig:overview_risk_guided_search}. We start with the specifications of safety-critical components and their uncertainty models. These models are used to study the effect of uncertainties on the occurrence of risks. Next, we apply grid-based IS to identify parameter regions that lead to large occurrence probabilities of dangerous events. Based on this, we can perform a risk assessment and draw conclusions about how a safe system design can be achieved. 
\subsection{Specification of Components and Uncertainties (Step 1)}
First, safety-critical system components and their uncertainty models must be specified. 
Usually, the uncertainties of technical devices are modeled by Gaussian distributions, where the mean value $\mu$ denotes the measured value and the standard deviation $\sigma$ the uncertainty.  
We refer to the manufacturer specifications (\eg data sheet) for the uncertainties of such components. 
In case of black box tools, where the uncertainty is unknown, we refer to our previous work \cite{Baek2023_IAS}, in which we showed how uncertainties of black box tools can be quantified.  
Briefly, this paper suggests to identify system properties that remain constant during operation (\eg positions of static objects, constant velocities,...) and can be measured by the black box tool of interest. After collecting data with the black box tool, its uncertainty is estimated via evaluating unexpected fluctuations regarding the constant properties. Thus, any deviations of constant system parameters in the measurement data are attributed to uncertainties of the employed tool.
%

\subsection{Experiments with Uncertainty Models (Step 2)}
To explore how spatial and temporal uncertainties contribute to the risk occurrence, we sample them from the respective probability distributions identified in the previous steps. Generally, we assume that the uncertainty models are specified by probability density functions. 
These samples are then accounted in the \emph{system dynamics} described by $w$ to obtain the system state $x(t)$, \ie
\begin{equation}
    \dot{x} = y(x(t),u_{a}). 
    \label{eq:state_equation}
\end{equation}
Here, $w$ may be given by an analytical function, specified manually for real-world systems or obtained via simulations. This means that the state variables in \autoref{eq:state_equation} and the uncertainties $u_a$ are specified individually for each system. In this paper, $w$ is provided by a simulator as we will elaborate on in \autoref{sec:experiments}. 
By evaluating the above equation for different values for $u_{a}$, where $a$ denotes the attribute as introduced in \autoref{subsec:MeasurementUncertainties}, we manipulate the system dynamics and observe how the uncertainties affect the risk occurrence. 


\subsection{Grid-based Importance Sampling (Step 3)} \label{GBIS}
The results from the previous step allow us to identify the interesting regions, that is, areas where the probability for dangerous events is large. 
However, these areas often suffer from low sample numbers leading to high VAEs. 
To overcome this, we apply grid-based IS as defined in \autoref{eq:GBIS}. 
In particular, our goal lies in exploring the suitability of this technique for risk assessment purposes. 
To do so, we perform the steps in the pseudocode in \autoref{euclid}. 
Specifically, the algorithm can be divided in two parts:
\begin{enumerate}
    \item In the \emph{learning phase} (lines 1 - 26 in \autoref{euclid}), classical MC sampling is performed in the entire sample space. Also, the sample space is partitioned into $G$ grids. After conducting the MC sampling and selecting a fraction of samples $\beta$ that are used for this learning phase, the clusters of critical samples are identified and sorted into the corresponding grids. Specifically, this is realized by means of a weighting function that assigns a weight of 1 to critical samples and 0 otherwise. To learn the density of critical samples, we divide the number of critical samples in each cell by the total amount of these. In addition, we add an absolute value for the Gaussian noise to avoid the effect of missing clusters. 
    \item To perform the importance sampling (lines 28-38 in \autoref{euclid}) , we take the remaining samples $n\cdot (1-\beta)$ and iterate over the grid cells. Again, we filter out critical samples with the above weighting process. By means of the density of each grid cell $g$ and the number of available samples $r$, we calculate the area of $g$. Afterward, this area is divided by the area of the entire sample space (line 33 in \autoref{euclid}). Finally, the samples of the grids are re-weighted.  
\end{enumerate}
In particular, the geometric area of each grid cell serves as the latent PDF $q(x)$ in \autoref{eq:GBIS}: Samples that occur in the cells with smaller areas are assigned to higher significance. It is worth noting that the area of grid cells may follow non-uniform density distributions. In that case, the respective density function must be considered.

\begin{figure}[t!]
	\centering
	\begin{subfigure}{.26\textwidth}
		\centering
        \includegraphics[width=\textwidth]{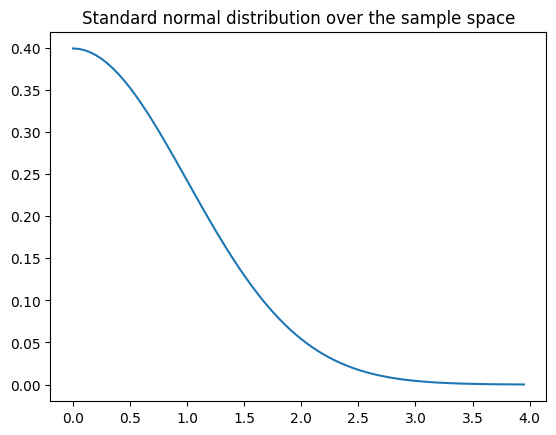}
		\caption{PDF of sample space \\ (standard normal distribution)}
		\label{subfig:IS_normal}
	\end{subfigure}%
	\begin{subfigure}{.24\textwidth}
		\centering
        \includegraphics[width=\textwidth]{Figures/IS_Grid.png}
		\caption{Respective grid cell densi-\\ties.}
		\label{subfig:IS_Grids}
	\end{subfigure}
	\caption{Grid densities obtained for the standard normal density distribution for the area of grild cells ($e=10$).}
	\label{fig:scenarios}
\end{figure}

\begin{algorithm} 
\caption{Grid-based Importance Sampling}\label{euclid}
\begin{algorithmic}[1]\label{alg:algorithm}
\Procedure{ImportanceSample}{}
\State $\textit{n} \gets \textit{number of total samples}$
\State $\beta \gets \textit{fraction of samples for learning phase s}$
\State $\textit{e} \gets \textit{number of edges per side}$
\State $\textit{Gaussian} \gets \textit{function to create Gaussian noise}$

\State 

\State $G \gets \text{Grid partition (sample space)}$
\State $\textit{S} \gets \text{sample(}\textit{n}*\beta\text{)}$
\State $\textit{C} \gets \text{Critical samples in S}$

\State Sort critical samples into partition \(G\):
\For {$c \in C$}
\State $i \gets \text{Index of cell that contains c in G}$
\State $\textit{G[i]} \gets \textit{G[i]} + 1$
\EndFor
\State $t \gets 0$ \Comment{Helper variable (density of all grid cells) }\\

\State Calculate total density (all cells):
\For {$g \in G$}
\State $g \gets \frac{g}{|C|}$
\State $g \gets g + \text{Gaussian()}$
\State $t \gets (t+g)$
\EndFor

\State Adjust density (account for added Gaussian noise):
\For {$g \in G$}
\State $g \gets \frac{g}{t}$
\EndFor

\State

\State $r \gets n*(1-\beta)$
\State $\text{output} \gets 0$

\For {$g \in G$}
\State $S \gets sample_g(r*g)$ \Comment{samples taken inside g}
\State $C \gets \text{critical samples in S}$
\State $A \gets \text{Area of g / Area of sample space}$
\State $\text{output} \gets |C| * \frac{A}{g}$

\EndFor

\State $\text{output} \gets \frac{\text{output}}{r}$
\State \textbf{return} output

\EndProcedure
\end{algorithmic}
\end{algorithm}

\subsection{Risk Reduction and Safe System Design (Step 4)} \label{subsec:step4}
After estimating the consequences of $u_s$ and $u_t$ on the risk probability and reducing the VAE via IS, we conduct a risk assessment by referring to \autoref{eq:risk}. 
Obviously, we are interested in identifying a component selection for 
\begin{equation}
    \min{risk} = \min{r(u_s,u_t)},  \; \; \mathrm{s.t.} \; \; Pr(r(u_s, u_t)) \leq \lambda
\end{equation}
for a safety limit $\lambda$. 
We thereby obtain the relationship $Pr(r(u_s, u_t))$. 
In fact, plotting this function, that is, the probability for the risk occurrence over the uncertainties, allows us to deduce whether the safety limits are met by the employed components. 
Particularly, this relationship provides the possibility to specify an uncertainty threshold and to select system components accordingly. 
We consider the VAE as a measure for the trustworthiness of our findings. 


\section{Experiments and Results}\label{sec:experiments}


\subsection{Three Scenarios}
\begin{figure*}[t!]
	\centering
	\begin{subfigure}{.33\textwidth}
		\centering
        \includegraphics[width=0.95\textwidth]{Figures/scenario1.png}
		\caption{Scenario A}
		\label{subfig:A}
	\end{subfigure}%
	\begin{subfigure}{.33\textwidth}
		\centering
        \includegraphics[width=0.95\textwidth]{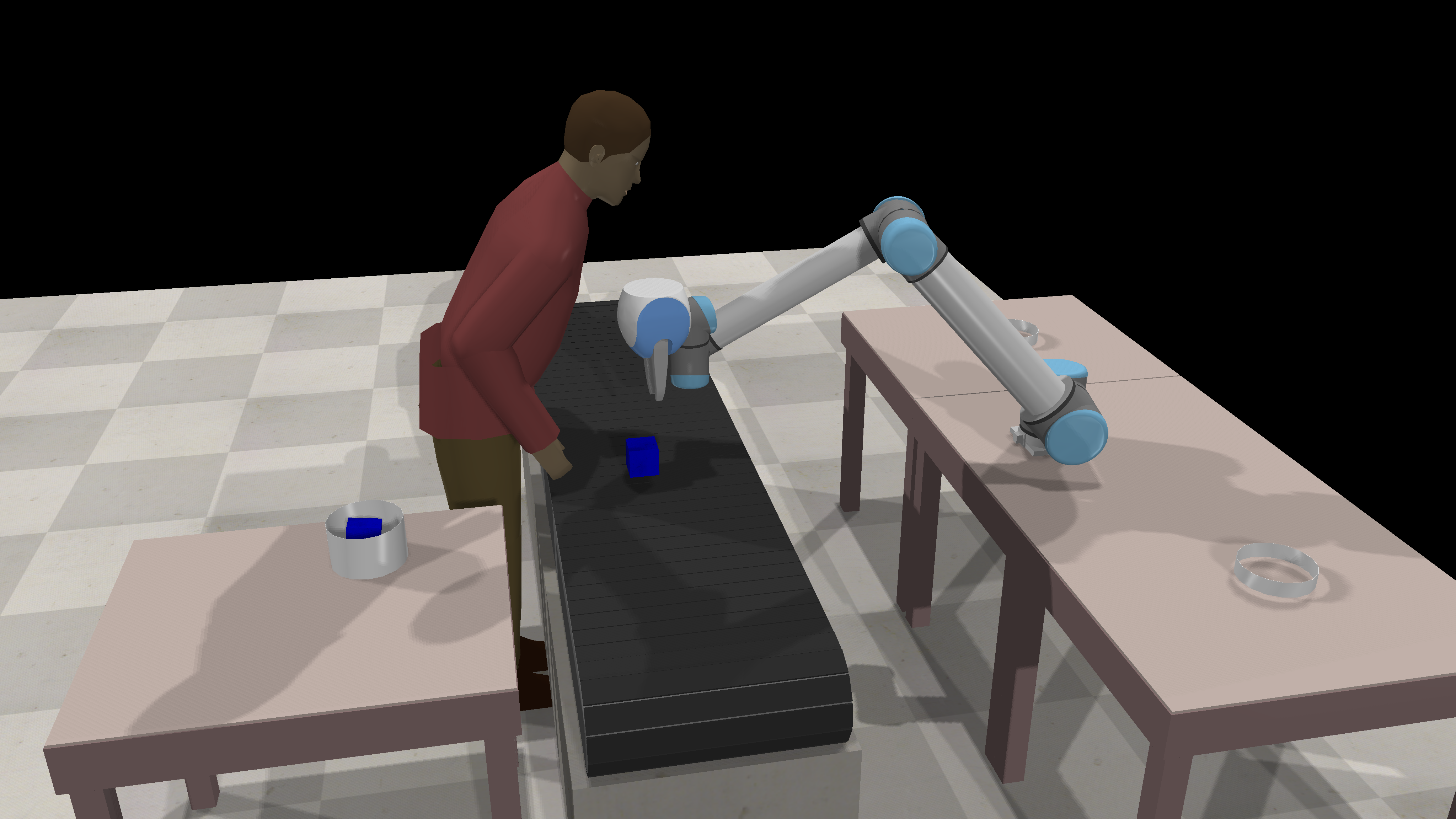}
		\caption{Scenario B}
		\label{subfig:B}
	\end{subfigure}
     \begin{subfigure}{.33\textwidth}
        \centering
        \includegraphics[width=0.95\textwidth]{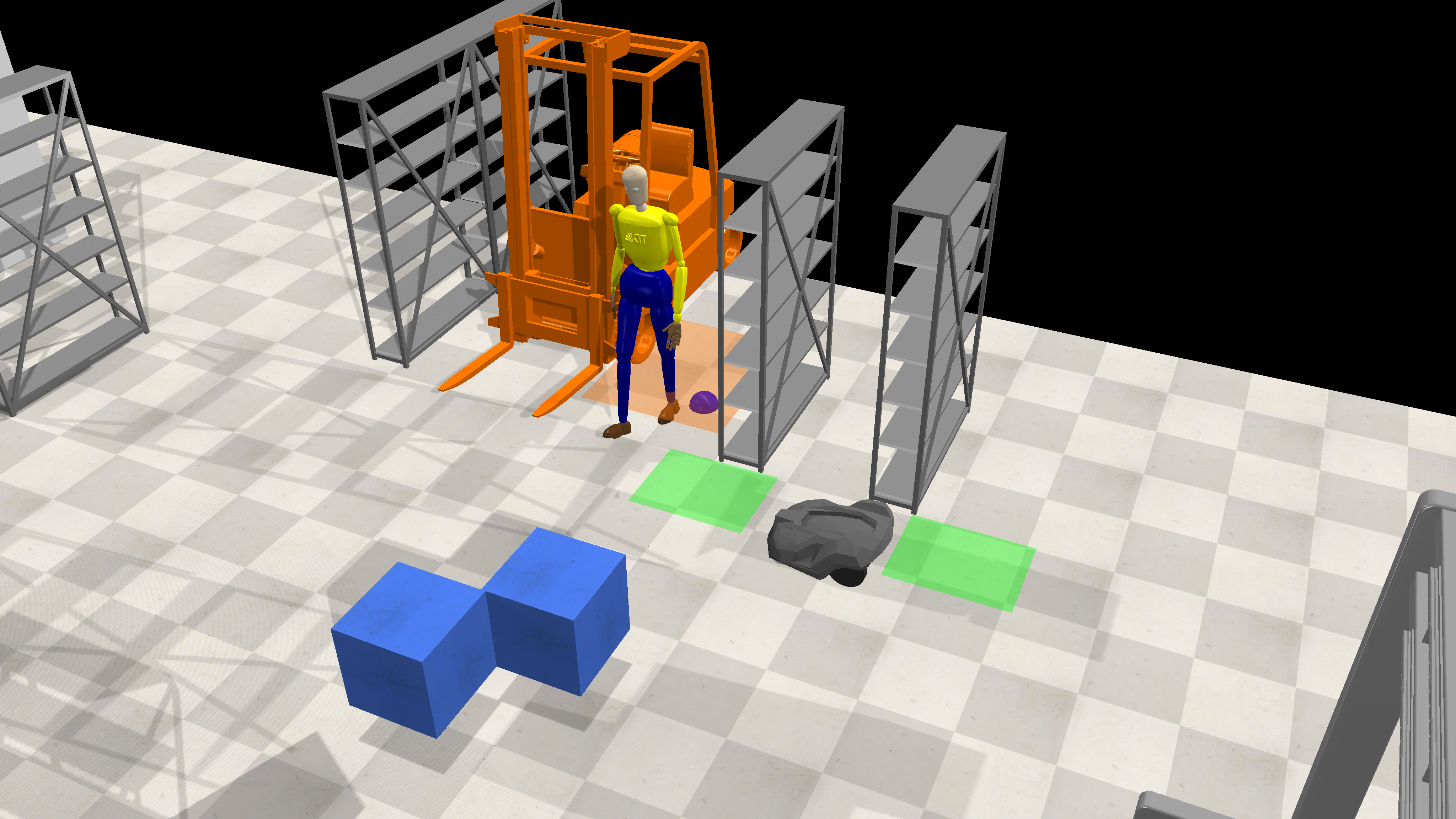}
        \caption{Scenario C}
        \label{subfig:C}
    \end{subfigure}
	\caption{We validate and evaluate our techniques by means of three HRC scenarios. In all scenarios, the robot performs evasive movements once the human-robot distance falls under a specified limit.} 
	\label{fig:scenarios}
\end{figure*}
We demonstrate the validity of above techniques in three HRC scenarios illustrated in \autoref{fig:scenarios}, where we investigate the impact of component-level uncertainties on system-level safety. Although the techniques are generalizable, we focus on measurement uncertainties since assessing the measurement accuracy is crucial for safe HRC. 
In our experiments, the system dynamics in \autoref{eq:state_equation} are obtained via simulations, where the continuous system dynamics are approximated by discrete values, \ie
\begin{equation}
    x_{k+1}= y_{HRC} (x_k, u_a) 
\end{equation}
for the time steps $k$. 
Importantly, we focus on safety-critical variables such as the positions and velocities of humans and robots. As will be specified in the following, we observe how manipulating these variables by the uncertainty samples $u_a$ contributes to the risk occurrence. 

\textit{Scenario A} (\autoref{subfig:A}): In this test scenario, a human approaches a robot arm which moves on a predetermined path. A distance sensor is mounted at the robot's end-effector to measure the human-robot distance $d_{HR}$. 
Once a certain distance threshold is met, the robot performs evasive movements to avoid collisions.

\textit{Scenario B} (\autoref{subfig:B}): The human and the robot reach for obstacles from a conveyor belt. Collisions may occur when both human and robot reach for the same object. 
In analogy to scenario A, the robot performs evasive movements if the distance falls below a specified threshold.

\textit{Scenario C} (\autoref{subfig:C}): A mobile robot navigates in a warehouse environment in the presence of a human. The mobile robot measures $d_{HR}$ using a front-mounted laser scanner and performs a safety stop if the distance is below a certain threshold.

\subsection{Parameter and Uncertainty Specification}
In all scenarios, the safety-critical parameters are subject to spatial and temporal uncertainties ${u_s, u_t}$ which are mo\-deled as follows:

\textit{Temporal uncertainties} are modeled by adding a time delay $\Delta t$ to the robot's safety reaction (\ie the evasive movement in scenarios A and B and the safety stop in scenario C). 
We consider a uniform distribution $\mathcal{U}\{0;9\}$ with $\mathcal{U}:\mathbb{N} \rightarrow \mathbb{R}$. 
We sample the uncertainty $u_t$ from this distribution. 
To be specific, the temporal uncertainty $u_t$ for one simulation run is given by:
$$u_t = N\cdot T ,$$
where $T$ is the simulation timestep and $N$ is sampled from $\mathcal{U}$.\\ 
\textit{Spatial uncertainties} are modeled by considering a spatial deviation in the distance measurement:
$$\tilde{d}_{HR}=d_{HR}+u_s, $$
where 
$d_{HR}$ is the ground-truth human-robot distance obtained from the simulator, $\tilde{d}_{HR}$ is the measurement value provided to the robot, and $u_s$ is the distance uncertainty. 
This distance uncertainty itself is calculated as  
$$u_d= \Delta d_0 + c\cdot v_{H}, $$
where $\Delta d_0$ is a constant value and $ c\cdot v_{H}$ is a velocity-dependant term with the human velocity $v_{H}$. This constant term is introduced because measurements of the human position are expected to underlie higher uncertainties for higher velocities. The constant value $\Delta d_0$ is sampled from a Gaussian while $c$ is sampled from a uniform distribution.


\subsection{Dangerous Zones}
Apart from studying the influence of uncertainties, we introduced an algorithm based on Grid-based IS in \autoref{GBIS}. 
The results from the previous step allow us to identify uncertainty constellations that lead to high risks.  
However, the sample number is relatively low in these regions which limits the statistical significance of additional analyses. 
To assess how \autoref{euclid} contributes to more reliable results, we investigate the detection performance regarding dange\-rous zones in above scenarios. 
In fact, we compare how our method based on IS performs in contrast to classical MC sampling.  

\subsection{Implementation and Execution of Experiments}
Our goal is to investigate how the aforementioned uncertainties contribute to the probability of human-robot collisions. To this end, the simulation scenarios are implemented in the \textit{CoppeliaSim} robotics simulator. The simulations are parameterized by drawing 10 samples for $\Delta t$, 10 samples for $c$, and 25 samples for $\Delta d_0$ from the respective distributions. This results in a total of 2500 parameter combinations for each scenario. To assess the influence of the uncertainties on the risk, a collision severity index is calculated in each simulation run:
$$
\mathrm{severity} = \begin{cases}
\label{eq:cases}
0\quad &\text{if no collision occurs}\\
\frac{F_c}{F_{max}}\quad &\text{in case of collisions}
\end{cases} 
$$
Here, $F_c$ denotes the estimated collision force and $F_{max}$ the respective limit for the affected body region as specified in \cite{ISO2016}. 
To evaluate whether the grid-based IS yields lower VAEs than MC sampling, we apply both techniques to identify the dangerous parameter regions.

\subsection{Results}\label{subsec:results}
\subsubsection{The Impact of Uncertainties on the Risk}\label{subsubsec:uncert}
\begin{figure*}[t!]
	\centering
	\begin{subfigure}{.33\textwidth} 
		\centering
        \includegraphics[width=0.95\textwidth]{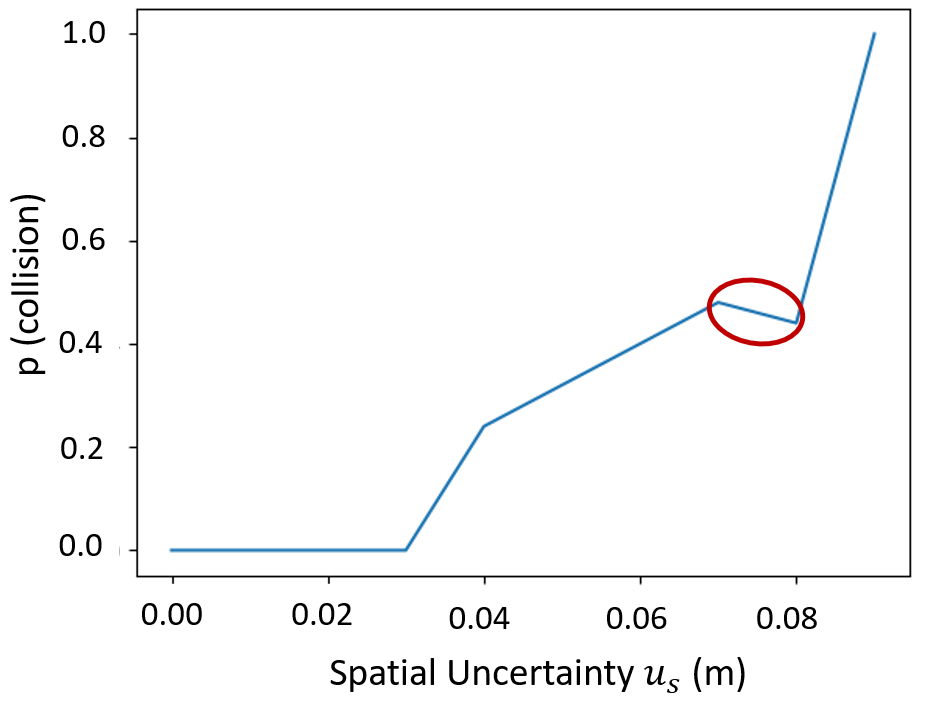}
		\caption{Collision probability vs. $u_s$}
  \label{subfig:spatial}
	\end{subfigure}%
	\begin{subfigure}{.33\textwidth}
		\centering
        \includegraphics[width=0.96\textwidth]{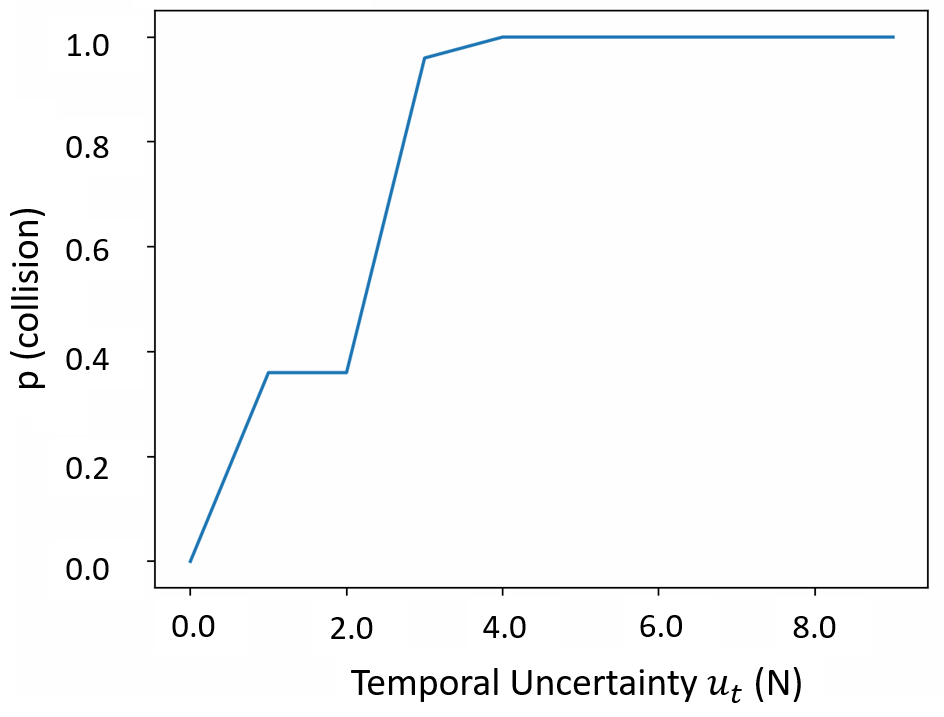}
		\caption{Collision probability vs. $u_t$}
  \label{subfig:temporal}
	\end{subfigure}
     \begin{subfigure}{.33\textwidth}
        \centering
        \includegraphics[width=\textwidth]{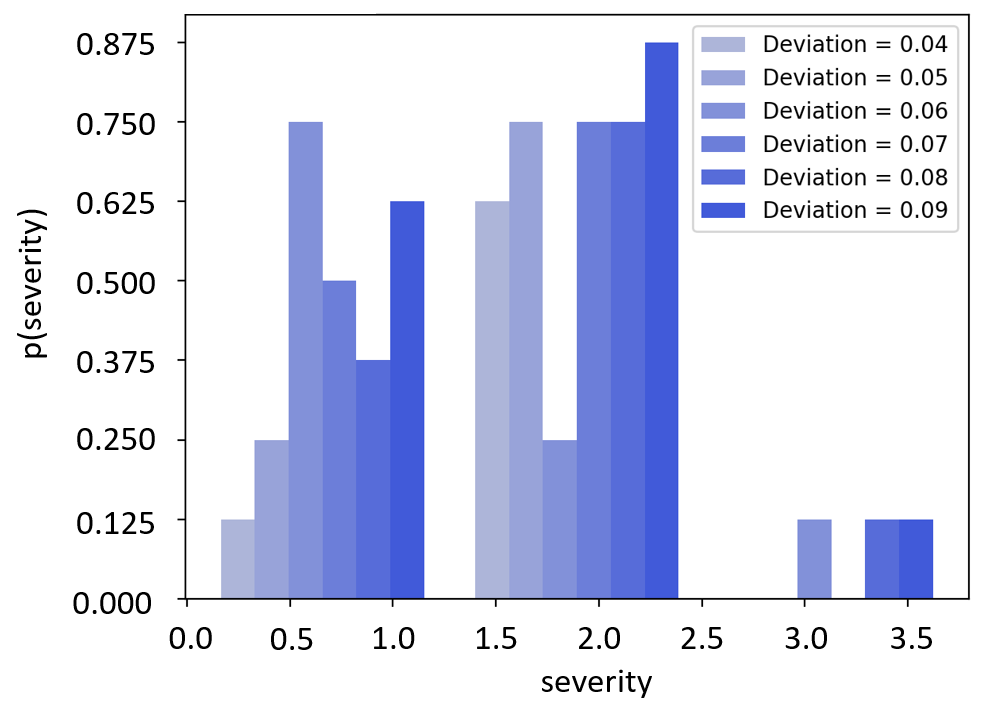}
        \caption{Occurrence of severity values}
        \label{subfig:hist}
    \end{subfigure}
	\caption{After studying how the collisions depend on the spatial and temporal uncertainties, we perform the risk evaluation by accounting for the collision forces.} 
	\label{fig:scenarios}
\end{figure*}

In order to explore the impact of $u_t$ and $u_s$ on the risk probability, we plot the number of events, where the distance falls below the specified distance threshold against the respective uncertainty. 
In doing so, we show the relationship between $u_t$, $u_s$ and the occurrence probability for dangerous events. 
As can be seen in \autoref{subfig:temporal}, dangerous events occur in 100\% once a temporal uncertainty of $u_t=4\,s$ is exceeded.  
Obviously, dangerous failures are more likely for higher uncertainties in general, which matches our expectations. 
However, depending on the considered application, certain constellations deliver unexpected results: For example in the region of the red circle in \autoref{subfig:spatial}, the collision probability seems to decrease for higher spatial uncertainties. 

\subsubsection{Importance Sampling vs. MC Sampling} \label{subsubsec_IS}
The obtained VAE values are shown in \autoref{tab:MC_IS}. 
While the VAE is decreased in scenario A and scenario C, an increase can be noted for scenario B. 
We collected 3000 samples for 10 runs, resulting in a number of 30.000 samples per scenario.

\begin{table}[b]
\caption{VAE values for MC Sampling and grid-based IS}
    \centering
    \begin{tabularx}{\columnwidth}{|c|c|c|>{\centering\arraybackslash}X|} \hline
        Sampling Method & Scenario A & Scenario B & Scenario C  \\ \hline
         MC Sampling    &  3.4 e-05 & 1.3 e-04 & 1.6 e-03  \\
        Grid-based IS & 2.8 e-05 & 3.8 e-04 & 7.4 e-04 \\ \hline
    \end{tabularx}
    \label{tab:MC_IS}
\end{table}


\def\Ha{
  \cellcolor{red!30} 
}
\def\Hb{
  \cellcolor{orange!80!red!25!white} 
}
\def\Hc{
  \cellcolor{yellow!35}
}
\def\Hd{
  \cellcolor{green!25}
}



\section{Data Analysis, Evaluation and Discussion}
We now conduct deeper analyses for above results by discussing the dependency of the risk occurrence probability on $u_t$ and $u_s$. 
Furthermore, we compare the performance of IS and MC sampling. 


\subsection{Relationship between Uncertainties and Risk}
The unexpectedly high collision probability at lower uncertainties described in \autoref{subsec:results} occurs due to the probabilistic nature of uncertainties: 
The samples in the red circle in \autoref{subfig:spatial} were generated in low uncertainty regions of $u_s$. 
This can be seen in \autoref{tab:corr}, that depicts the influence of uncertainties on the collision probability.  
For example, the collision probability for $u_s=0.3\,m$ and $u_t=0.09\,s$ yields a lower value than for $u_s=0.1\,m$ and $u_t=0.08\,s$.  
Since in contrast to errors, uncertainties are modeled distributions, we conclude that low values for $u_s$ and $u_t$ were sampled in these cases. 
We thereby infer that deriving a generally valid analytical function is not possible, which motivates to apply our sampling-based method and to analyze the results individually for each system. 

\subsection{VAE Reduction via Importance Sampling}
Apart from the dependency between the evolvement of dangerous events and measurement uncertainties, the occurrence probability of these situations is often very low. 
As \autoref{tab:MC_IS} shows, the VAE decreases when applying the grid-based IS instead of MC sampling for scenarios A and C. 
However, in scenario 2, the higher grid resolution yields a higher VAE which can be explained as follows:
Here, the dangerous zone is relatively large such that the assumption of a low probability density for the interesting region is not valid. 
Consequently, the learning phase of the grid-based IS is subject to an underestimation of the dangerous zone, that significantly affects the VAE. 
In this case, MC sampling outperforms our approach. 

\subsection{Uncertainty-aware Safety Evaluation}
To evaluate safety limits and derive suggestions for the design of safe systems, we plot the probability for risks (collisions) calculated via \autoref{eq:risk} and \autoref{eq:cases} over the uncertainty for all experiments. 
Based on these, we obtain the distribution for the severity shown in \autoref{subfig:hist} that allows to directly evaluate arbitrary safety limits: 
By comparing the thresholds for the risk probability and the severity with our results, we are able to deduce the tolerated uncertainty as explained in \autoref{subsec:step4}. 
Finally, this uncertainty threshold enables us to identify components (\eg sensors, cameras,...) that are needed to design safe systems.


\subsection{Limitations}
While our results demonstrate the validity and performance of our novel method for the uncertainty-aware risk assessment, we note that further studies are necessary for a thorough evaluation. 
To be specific, we considered three HRC scenarios modeled in \emph{CoppeliaSim}. Although we co\-vered the case of an industrial robot system (scenario B) and an example with a mobile robot (scenario C), investigating a larger variety of robot systems is required. 
In addition, we introduced grid-based IS to enhance the significance of our results. 
We found that the performance strongly depends on the parameter selection. 
To enable the applicability of our method to more complex scenarios, we note that a method to optimizing the parameters is necessary. 

\begin{table}[h!!]
  \caption{Collision probability for different uncertainty constellations ($u_s$ and $u_t$).}
  \begin{adjustbox}{width=\columnwidth,center}
    \begin{tabular}{|c| c | c | c | c | c | c | c | c | c | c |}
      \hline
      \diagbox[width=8em]{Time\\Delay}{Sensor\\Deviation\\$\times10^{4}$}& \textbf{0}&\textbf{5}&\textbf{10}&\textbf{15}&\textbf{20}&\textbf{25}&\textbf{30}&\textbf{35}&\textbf{40}&\textbf{45}\\
      \hline
      \textbf{0} & {N/A}       & {N/A}       & {N/A}      & {N/A}      & \Ha{0.599} & \Hb{0.861} & \Hb{0.691} & \Hb{0.791} & \Hc{0.858} & \Hb{0.63}  \\
      \textbf{1} & {N/A}       & {N/A}       & {N/A}      & \Ha{0.069} & \Hb{0.704} & \Ha{0.14}  & \Hb{0.721} & \Hb{0.721} & \Hc{0.816} & \Hd{0.917} \\
      \textbf{2} & {N/A}       & {N/A}       & {N/A}      & \Ha{0.424} & \Ha{0.472} & \Hc{0.877} & \Hb{0.705} & \Hc{0.841} & \Hb{0.731} & \Hb{0.629} \\
      \textbf{3} & {N/A}       & {N/A}       & \Hb{0.612} & \Hb{0.535} & \Hc{0.819} & \Hd{0.899} & \Hb{0.798} & \Hb{0.745} & \Hc{0.838} & \Hb{0.768} \\
      \textbf{4} & {N/A}       & {N/A}       & \Hb{0.615} & \Hb{0.671} & \Hb{0.712} & \Hc{0.874} & \Hb{0.738} & \Hc{0.84}  & \Hb{0.738} & \Hd{0.9}   \\
      \textbf{5} & {N/A}       & \Hb{-0.611} & \Ha{0.359} & \Hc{0.823} & \Hd{0.914} & \Hb{0.757} & \Hd{0.894} & \Hb{0.626} & \Hc{0.819} & \Hc{0.878} \\
      \textbf{6} & \Hb{-0.374} & \Hb{-0.787} & \Ha{0.216} & \Hb{0.737} & \Hc{0.862} & \Hd{0.954} & \Hd{0.886} & \Hd{0.876} & \Hb{0.75}  & \Hd{0.941} \\
      \textbf{7} & \Hb{-0.799} & \Ha{-0.615} & \Ha{0.35}  & \Hb{0.745} & \Hb{0.819} & \Hb{0.681} & \Hc{0.813} & \Hd{0.954} & \Hc{0.778} & \Hc{0.878} \\
      \textbf{8} & \Hb{-0.803} & \Ha{-0.402} & \Ha{0.167} & \Ha{0.511} & \Hc{0.813} & \Hb{0.754} & \Hb{0.75}  & \Hc{0.87}  & \Hd{0.904} & \Hd{0.912} \\
      \textbf{9} & \Hb{-0.379} & \Ha{0.085}  & \Hb{0.433} & \Ha{0.54}  & \Hb{0.522} & \Hc{0.839} & \Hb{0.505} & \Hb{0.714} & \Ha{0.393} & \Hb{0.681} \\
      \hline
    \end{tabular}
  \end{adjustbox}
  \label{tab:corr}
\end{table}

\section{Conclusion and Outlook}
In this work, we derived a technique to evaluate risks of robot systems in an uncertainty-aware manner. 
We presented the grid-based importance sampling that enables to draw inferences with enhanced statistical significance for regions underlying sparse probability densities as for example dangerous events in HRC. 
We validated and applied our methods on three HRC scenarios modeled in the simulation environment \emph{CoppeliaSim}. 
In doing so, we explored how spatial and temporal measurement uncertainties impact the risk probability. 
Furthermore, we showed how our results can be used to evaluate arbitrary safety limits that define thresholds for the risk probability and the severity of accidents. 
In the future, we plan to study the generalizability of our methods by exploring its applicability to real-world HRC scenarios.









\bibliographystyle{IEEEtran}

\bibliography{references}




\end{document}